\documentclass{egpubl}
\usepackage{eurovis2020}

\SpecialIssuePaper         

\usepackage[utf8]{inputenc}

\usepackage[T1]{fontenc}
\usepackage{dfadobe}  

\usepackage{cite}  
\BibtexOrBiblatex
\electronicVersion
\PrintedOrElectronic
\ifpdf \usepackage[pdftex]{graphicx} \pdfcompresslevel=9
\else \usepackage[dvips]{graphicx} \fi

\usepackage{egweblnk}
\usepackage{hyperref}
\usepackage{amsmath}
\usepackage{cleveref}
\title[Classifier-Guided Visual Correction]%
      {Classifier-Guided Visual Correction\\of Noisy Labels for Image Classification Tasks}

\author[A. Bäuerle, H. Neumann, and T. Ropinski]
{\parbox{\textwidth}{\centering
        A. Bäuerle \orcid{0000-0003-3886-8799},
        H. Neumann \orcid{0000-0001-7687-5792},
        and T. Ropinski \orcid{0000-0002-7857-5512} 
        }
        \\
{\parbox{\textwidth}{\centering 
        All authors are with Ulm University. E-mail: {alex.baeuerle|heiko.neumann|timo.ropinski}@uni-ulm.de.
       }
}
}

\begin{document}

\maketitle
\begin{abstract}
Training data plays an essential role in modern applications of machine learning.
However, gathering labeled training data is time-consuming.
Therefore, labeling is often outsourced to less experienced users, or completely automated.
This can introduce errors, which compromise valuable training data, and lead to suboptimal training results.
We thus propose a novel approach that uses the power of pretrained classifiers to visually guide users to noisy labels, and let them interactively check error candidates, to iteratively improve the training data set.
To systematically investigate training data, we propose a categorization of labeling errors into three different types, based on an analysis of potential pitfalls in label acquisition processes.
For each of these types, we present approaches to detect, reason about, and resolve error candidates, as we propose measures and visual guidance techniques to support machine learning users.
Our approach has been used to spot errors in well-known machine learning benchmark data sets, and we tested its usability during a user evaluation.
While initially developed for images, the techniques presented in this paper are independent of the classification algorithm, and can also be extended to many other types of training data.
\\


\end{abstract}  

\section{Introduction}
While most of the latest breakthroughs in deep learning have been achieved by means of supervised algorithms, these algorithms have one essential limitation: they require large amounts of labeled training data.
When learning image classification tasks, this means that a large set of correctly labeled images needs to be available~\cite{nettleton2010study,pechenizkiy2006class}.
Since the labeling process is time-consuming and labor-intensive, acquiring labeled training data is, however, a cumbersome process.
To speed this process up, labeling is often outsourced to less experienced annotators or crowd workers, for instance via Amazon's Mechanical Turk~\cite{khetan2017learning,rashtchian2010collecting}.
In the context of deep learning, crowd workers are human labor, getting paid for labeling large data sets.
Sometimes, even automatic label assignment tools are used~\cite{usami2011automatic}.
Unfortunately, such a label acquisition process usually leads to noisy labels, i.e., a training data set which contains many wrongly assigned labels.
This can compromise training results~\cite{zhang2016understanding}.
Thus, to be able to benefit from these approaches for training data acquisition, dedicated quality control mechanisms must be in place.

\begin{figure}[t]
  \centering
  \includegraphics[height=4.5cm]{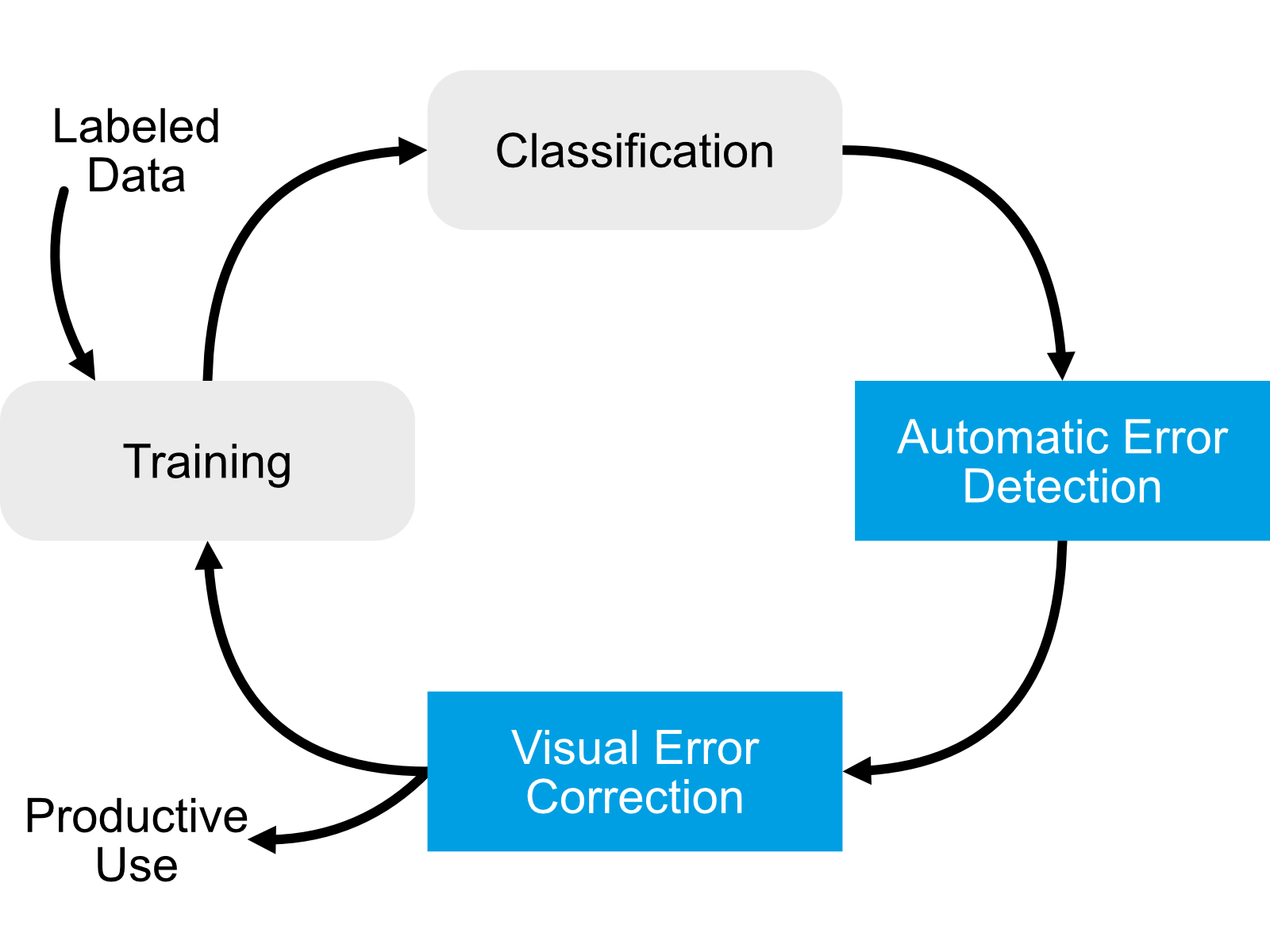}
  \caption{
    We propose a classifier-guided \textbf{Automatic Error Detection} for noisy labels, to visually guide the user to erroneous labels, which can then be inspected and corrected during the proposed \textbf{Visual Error Correction}.
    The two proposed components seamlessly integrate with standard machine learning workflows, as they operate downstream from \textbf{Training} and \textbf{Classification}.
    After a visual inspection, the classifier can be deployed for \textbf{Productive Use}, or trained again to be iteratively improved through the proposed process.
  \label{fig::process_new}}
\end{figure}

To address the problem of noisy labels, we propose a classifier-guided visual correction approach, which combines automatic error detection with interactive visual error correction (see~\Cref{fig::process_new}).
To enable the automatic detection, we have systematically categorized error types, that can be potentially present in noisy label data sets.
Our categorization led to three such error types: \emph{Class Interpretation Errors}, \emph{Instance Interpretation Errors}, and \emph{Similarity Errors}.
Tailored towards these error types, we further introduce detection measures, which are based on the classifier's response.
Therefore, we first train with the potentially noisy labels, and subsequently classify all training and validation images with the trained classifier.
The classifier's response can then be analyzed using our error detection measures to guide the user to potential errors.
These potential errors are visualized in a way that supports an interactive visual correction.
To visually guide the user during the correction, we propose to employ linked list visualizations with importance sorting.
By using our approach, the number of required inspections is bound by -- and usually much lower than -- the classification error, i.e., for a classifier that reaches an accuracy of 92\%, only 8\% of the data has to be reviewed at maximum.
While this is the upper bound for investigated training samples per iteration, all samples that have already been inspected can additionally be ignored in the error detection process in future iterations.
This means that for each subsequent iteration of data-cleanup, only those samples where the classifier disagrees with the label and that have not been already revisited need to be reviewed.
Without our classifier-guided approach, instead, an inspection of the entire labeled data set would be necessary.

As illustrated in~\Cref{fig::process_new}, the proposed approach can be used iteratively to further improve classification accuracy, whereas users have to inspect fewer images for each subsequent iteration, as already inspected images do not require further consideration.
While the contributions made in this paper address automatic error detection and visual error correction, no modifications are necessary for collecting labels, or training and testing the classifier, as our approach is to correct training data independent of the labeling or training process, allowing data experts to review data sets that have been fully labeled.
This is in contrast to active learning, which modifies the label acquisition process during training~\cite{seung1992query,settles2010active}, as well as more recent fully automatic techniques, which modify the training process, and also reduce the amount of training data by sorting out noisy labels~\cite{tanaka2018joint,lee2018cleannet,huang2019o2u}.
We propose an error correction approach that is based solely on classification results of the trained model, and integrates seamlessly with modern deep learning workflows without reducing the size of the training data set.

\noindent To this end, we make the following contributions throughout this paper:

\begin{itemize}
    \item Categorization of label error types potentially occurring in classification training data.
    \item Measures to identify error candidates by means of classifier response analysis.
    \item Interactive visual error guidance and correction by means of classifier result measure visualization.
\end{itemize}

\begin{figure}[b]
  \centering
  \includegraphics[width=0.8\columnwidth]{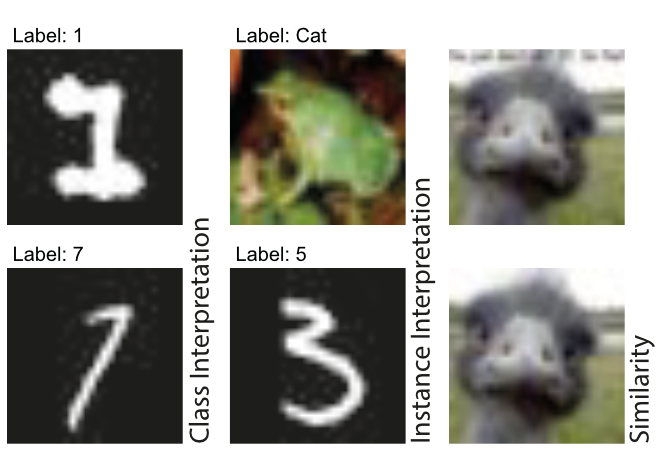}\hfill%
  \caption{
  Examples of errors we discovered by applying our techniques to widely used machine learning benchmark data sets.
  On the left, one can see possible Class Interpretation Errors.
  While the top one was labeled as one, the bottom one was labeled to be a seven.
  The frog in the center is labeled as a cat and the three as a five, thus, single instances were clearly misinterpreted.
  On the right, one can see almost equal images.
  One might question if they should both be in the data set.
  (Original resolution of 32 by 32 for Cifar10/animals and 28 by 28 for MNIST/digits)\label{fig::found}}
\end{figure}

We have realized these contributions within an interactive visualization system, with which we were able to identify errors in standard machine learning benchmark data sets, such as MNIST and CIFAR10 (see~\Cref{fig::found}).
We have further evaluated this system, whereby our findings indicate, that it enables users to intuitively clean noisy label data in order to achieve higher classification accuracies.

\section{Related Work}

Work on handling noisy labels for datasets can be delineated into two main categories.
On one side, some approaches aim at inspecting datasets, often through visualization.
On the other, there are training setups that aim at providing robust classifiers that cope with noisy labels.
The following will provide an overview of both those lines of research.

\noindent\textbf{Data labeling.}
One area of deep learning where data labeling is a central aspect is active learning~\cite{settles2010active}.
Here, candidates for labeling assignments are selected, often through a query-by-committee strategy, where the output of several classifiers is used to inform candidate selection~\cite{seung1992query}.
The line of work by Bernard et. al.~\cite{bernard2017comparing,bernard2018towards,bernard2019visual} investigates how label acquisition in active learning scenarios can be improved.
They also employ the classifier directly to suggest new items to be labeled and use dimensionality reduction techniques to visualize these proposed items and their distribution.
What separates active learning from our work is, that active learning does not aim at improving noisy data sets, but rather works towards improving the labeling process itself.
Thus, active learning is placed before label acquisition has been performed, while our approach is designed to work with readily labeled data sets.

There also exist numerous techniques to ensure a better quality of crowdsourced training data while labels are being generated~\cite{heimerl2012visual,chang2017revolt,settles2011closing}.
They use multiple workers~\cite{kairam2016parting}, provide monetary benefits for good work and specialized task framing~\cite{rogstadius2011assessment}, or select workers with predefined requirements~\cite{mitra2015comparing}.
All of these approaches are focused on quality assurance while labels are acquired.
Approaches to examining data quality after labeling through crowd services are analyzing how the worker interacted with the system~\cite{rzeszotarski2012crowdscape,liu2019interactive}, or having workers review the work of other workers~\cite{hansen2013quality}.
A work published by Chang et. al. combines multiple of these aspects to ensure data quality by grouping multiple workers and letting them interactively reason about label decisions~\cite{chang2017revolt}.
However, they do not incorporate the classifier feedback in their visualizations, which is the building block of our guidance system and can help reduce the samples to be revisited.
Additionally, their techniques are only applicable if all annotations are present and can be assigned to individual workers.
Yet, correcting labels for readily provided data sets where original labelers are not accessible anymore can be valuable to support already processed data sets.
Current tools are targeted more towards analyzing worker performance than correcting already labeled data sets.
However, it can often be of great value for domain experts to be able to validate and correct their training data themselves as sometimes the data is specific and cannot be perfectly labeled by laymen.
Additionally, for all of these data improvement methods in the context of crowdsourcing, one needs to either hire more crowdworkers, refine the requirements or conduct a separate, second task to verify the generated labels, which comes with a greater investment of money and time during label acquisition~\cite{sheng2008get, ipeirotis2014repeated}, and sometimes even makes crowdsourcing more expensive than conventional approaches~\cite{khetan2017learning}.
In these scenarios, it is therefore helpful if domain experts can review and resolve label errors quickly.
Our approach is thus focused on correcting erroneous labels.

Visualization has been used for data cleaning in several publications, which shows how effective visualization can be when data is to be cleaned.
Kandel et. al. worked on improving data by visually exploring data sets and directly manipulating them whenever a user spots a problem in the data~\cite{kandel2011wrangler,kandel2012profiler}.
Gschwandtner at. al.~\cite{gschwandtner2014timecleanser} as well as Arbesser et. al.~\cite{arbesser2016visplause} use visualization to clean up time-oriented data.
Wilkinson developed measures and visualizations to detect and inspect outliers in data sets~\cite{wilkinson2017visualizing}.
However, these and related~\cite{park2016c,willett2013identifying} tools are not tailored towards use with machine learning data sets, which often exceed the amount of data used in these contexts, contain labels that are to be corrected instead of direct data properties and offer additional guidance usable for visualization designs, such as classification results.

In a publication by Xiang et. al., visualization is directly used to improve the quality of neural network training data sets~\cite{xiang2019interactive}.
They use a projection of the whole high dimensional data set to define trusted items, which are then propagated to more items using an approach by Zhang et. al.~\cite{zhang2018training}.
However, while this approach combines human interaction with network-based label correction, they do not use the network predictions as guidance to potential errors.
Similarly, Alsallakh et. al.~\cite{bilal2018convolutional} developed a visualization method to analyze class hierarchies in training scenarios.
The purpose of this approach is to identify class hierarchies that are often confused by the classifier, and upon this knowledge, improve the classifier or label definitions.
As a side-product, they were also able to find labeling errors in the data.
However, their visualization design and especially the lack of tools to directly investigate and correct mislabeled samples shows, that this is not the main goal of their application.

\noindent\textbf{Robust training.} One way to approach noisy data sets is to train a classifier that is robust against such noisy labels.
Here, some approaches rely on modifications of said classifier to introduce features that can filter noisy labels~\cite{tanaka2018joint,zhang2018generalized,huang2019o2u}.
This introduces additional overhead and does not improve the general label quality so that the data set remains erroneous.
Others rely on additional, clean data to filter for noisy samples~\cite{patrini2017making,hendrycks2018using}.
These methods remove potentially noisy labels from the data set entirely~\cite{nguyen2019robust}, or reduce the importance of potentially false labels for training~\cite{reed2014training, jiang2017mentornet, ren2018learning}, which might reduce diversity in the data set.
Such approaches can help circumvent some of the downsides of data sets that contain labeling errors, however, they do not tackle the underlying problem.
Cleaning up data sets is still fundamental, as this is the only way a data set can be reliably reused, shared and published.
At the same time, these approaches effectively make the data set smaller, which is not desirable.
Some of these approaches also require using adjusted classifiers, which is neither desirable nor easy to use, especially by data-experts who are less experienced in ML.

Other authors introduce additional label cleaning networks to be trained to remove or relabel potentially compromised samples~\cite{veit2017learning,lee2018cleannet}.
Han et. al. even propose to use a self learning approach to clean up noisy labels using extracted features from the data points~\cite{han2019selflearning}, however, all these automatic approaches do not guarantee correct labels.
They either reduce the data set size, require modified training with another classifier, or both.
Additionally, they do not allow data-experts to verify their data sets.

We propose an approach to improve the training data set without having to look at every individual sample by using the classifier as a guide to mislabeled samples.
Our user-centered approach does not only focus on the final classifier performance, but is also targeted at cleaning up the training data at the same time, as it does not simply reweight or remove training samples. 
As this permanently corrects training data, it additionally makes the data reusable, publishable, and shareable.
Also, the approach we propose can directly be integrated into any training process, as it does not require any manipulation of the classifier or additional data.
Users simply use their trained classifier for permanent data-cleanup.
It additionally provides insights about the training data, e.g. which classes are typically confused, biased, or seen similar.

\section{Automatic Error Detection}\label{sec::types}

To be able to tailor the visual user guidance towards relevant errors in labeled data sets, a characterization is required to differentiate error types potentially occurring in such labeling scenarios.
Based on a systematic analysis of the image labeling process, we have identified three such error types.

Whenever annotators assign an incorrect label to an image, this can stem from two fundamentally different problems.
Either, they just mislabel the one image at hand, while they have in general understood the task; or they have a wrong mental image of a class, and thus assign incorrect labels to all data points of that class.
While these are problems that occur during the labeling of data points, another source for corrupted data sets may already be the data acquisition process.
Similar or equal data points are sometimes added to the data set more than once, which can shift the data-distribution away from real-world scenarios.
While the aforementioned error-types mostly stem from human errors, the addition of highly similar data points can be a problem especially when automatically collecting data, e.g. from online sources.
To summarize, noise in training data can be introduced when:

\begin{enumerate}
  \item A labeler confuses two classes (Class Interpretation Error)
  \item A labeler mislabels one data point (Instance Interpretation Error)
  \item Data points get added to the data set multiple times (Similarity Error)
\end{enumerate}

These error types all introduce unique challenges for how to resolve them.
Nevertheless, this categorization also enables the invention of novel approaches, to guide the user to potential instances of these errors.
Therefore, to suggest further inspection of labeled data points, we propose the following measures for the three error types:

\begin{enumerate}
  \item Class Interpretation Error Score (\Cref{eq:amount})
  \item Instance Interpretation Error Score (\Cref{eq:score})
  \item Similarity Error Score (\Cref{eq:similarity})
\end{enumerate}

For the first two scores, we use the classification results in combination with the labels, which might be incorrect, as the basis for computing them.
The Class Interpretation Error Score is computed for each label/classification (lbl/cls) subset of the data, whereas the Instance Interpretation Error Score is computed on individual instances.
The Similarity Error Score is computed for each instance pair with the same classification.
We assume that, although the labeled data may contain errors, the convolutional neural network (CNN) is still able to differentiate between different classes, such that in general incorrectly labeled data points get classified into their original class.
This assumption has been tested on an intentionally corrupted data set, which is described in~\Cref{sec::evaluation}.
Since this makes the classification result and the original label differ, these data points can be detected by looking at misclassifications in the data set.
The similarity error score instead, can be calculated by exploiting similarity measures between training samples.
As every part of the data-split can contain errors, we classify all samples in the data set once after the network has been trained.
This includes train, test, and validation data, which can then subsequently be corrected.
In the following, we introduce these three scores and their computation in detail.

\subsection{Class Interpretation Errors}

\begin{figure}[t]
  \centering
  \includegraphics[width=.14\columnwidth]{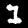}
  \includegraphics[width=.14\columnwidth]{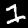}
  \includegraphics[width=.14\columnwidth]{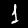}
  \includegraphics[width=.14\columnwidth]{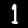}
  \includegraphics[width=.14\columnwidth]{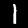}
  \includegraphics[width=.14\columnwidth]{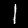}
  \includegraphics[width=.14\columnwidth]{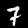}
  \includegraphics[width=.14\columnwidth]{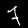}
  \includegraphics[width=.14\columnwidth]{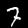}
  \includegraphics[width=.14\columnwidth]{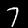}
  \includegraphics[width=.14\columnwidth]{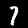}
  \includegraphics[width=.14\columnwidth]{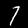}\mbox{}
  \caption{Images from the original MNIST data set (original resolution 28 by 28).
  The top row shows images labeled as one.
  The bottom row contains images labeled as seven.
  Here, Class Interpretation Errors might occur, since those digits are written differently in the US and Europe.\label{fig:sevenone}}
\end{figure}

Class Interpretation Errors are introduced when data points from class $a$ were assumed to be of class $b$ by one, or few, of the labelers.
This error type is conceptual, and leads to multiple or all labels assigned by one, or a few, labelers and belonging to class $a$ ending up with the wrong label $b$ (e.g., labelers considering gooses to be ducks throughout the entire data set).
However, as long as the majority of data points are correctly labeled, our presented approach is able to guide to these errors, as the classifier will still be able to correctly classify most of the data points with incorrect labels, see~\Cref{sec::evaluation}.
Fortunately, the fact that multiple data points are labeled incorrectly makes Class Interpretation Errors easy to detect.
We make use of the amount of resulting misclassifications to find candidates for Class Interpretation Errors. Thus, we analyze lbl/cls combinations by the amount of missclassifications in them as:

\begin{equation}\label{eq:amount}
CIES_{y,\hat y} = \vert \{x \mid x \in D, argmax(cls(x)) = y, lbl(x) = \hat y \} \vert
\end{equation}

Which means that the Class Interpretation Error Score $CIES$ given a prediction class $y$ and a ground truth class $\hat y$ is defined as the cardinality of the subset of data points $x$ in the data set $D$ for which the classification result $cls(x)$ equals $y$ and the label $lbl(x)$ equals $\hat y$.
Thus, this measure is designed to analyze entire lbl/cls subsets of the data.
An interesting occurrence of this type of error in the widely used MNIST data set is the interpretation of the North-American and European way of writing the digits '7' and '1', as shown in~\Cref{fig:sevenone}.

\subsection{Instance Interpretation Errors}

When single items in the data set get labeled incorrectly, the situation is more difficult, as these errors cannot be spotted by analyzing the ratio of misclassifications of one lbl/cls pair.
At the same time, however, they have less influence on classification accuracy as compared to Class Interpretation Errors.
To provide means to identify and remove Instance Interpretation Errors, we employ the classification confidence as an indication for labeling errors.
This works well for all probabilistic models, such as neural networks, where prediction probabilities are an implicit output.
When data points are misclassified confidently, they might as well be incorrectly labeled. This can be used to guide the user to these samples in the data set.
To enhance this guidance, we go one step further and analyze the relation of the classification confidence and the classification probability assigned to the ground-truth label of a data point.
On these means, Alsallakh et. al~\cite{bilal2018convolutional} state:

\emph{[...] detecting mislabeled samples such as an image of a lion labeled as a monkey.
We found such cases by inspecting misclassified samples having high prediction probability and low probability assigned to the ground-truth.}

\noindent We, therefore, propose the following measure to guide users to these error types:

\begin{equation}\label{eq:score}
IIES_x = \frac{\max(cls(x)) + (1 - cls(x)_{\hat y})}{2}
\end{equation}

Here, we calculate the Instance Interpretation Error Score $IIES$ for a data point $x$ as the normalized relation between the class that the classifier assigned the highest classification probability to, and the probability the classifier assigned to the ground-truth label $\hat y$.
Thus, this score provides guidance on an individual instance level.
This score is used as an indicator for how certain the classifier is wrt.~the misclassification of a data point, and can be used to recognize potential labeling errors.
Applying this approach to the widely used Cifar10 as well as the MNIST data set, revealed previously unknown labeling errors, which we discuss in~\Cref{sec::evaluation}.

\subsection{Similarity Errors}

When data points occur more than once in the labeled data set, this can lead to an unintended shift away from the real-world data distribution.
Such errors can be introduced when data points are taken from online sources or when an overview of the data set is not always present during data acquisition.
It is important to differentiate between intentionally augmented data and data points that might over-represent certain features during training, as data-augmentation can lead to better training results.
However, having multiple similar data points unintentionally in the labeled data set can compromise the training results in multiple ways.
When they are in the training set, a higher priority is assigned to this representation, which can lead to bias, where some features are considered more important than other features.
This is a problem when this over-representation is not expected in the productive use of the classifier.
When, in contrast, several instances are in the validation data set, validation accuracy has a higher variation depending on the correctness of the classification of these data points, which in turn might compromise the performance measure of the classifier.
If similar data points exist across training and validation data sets, validation is performed on data points that the classifier has been trained on, which can also compromise validation results, and at the same time introduce bias to the training data.
Gladly, guiding users to similar data points is also possible, as similarity measures can be computed for each pair of elements in the data set that are assigned the same classification result:

\begin{equation}\label{eq:similarity}
  \begin{split}
    & SES_{x_1,x_2} = sim(x_1, x_2),\quad for\ x_1, x_2 \in M \\
    & M := \{x_1, x_2 \in D \mid x_1 \neq x_2, argmax(cls(x_1)) = argmax(cls(x_2))\}
  \end{split}
\end{equation}

The Similarity Error Score $SES$ for a pair of data points $x_1, x_2$ can be obtained using similarity measures, which exist for many types of data.
The $SES$ is calculated for all pairs of data points in the data set $D$ that were classified into the same class, whereas the $sim$ function represents a similarity measure for two data points.
For images, this function could be the Structural Similarity Index Measure (SSIM)~\cite{wang2004image}.
While proposing candidates with this measure is not complex, Similarity Errors require the most experience of all error types to be resolved, as highly similar images are not always a problem for training a classifier. They are only harmful if either, they do not represent the real-world distribution, or, if they originate from both the training and validation data sets because then, validation does not test generalizability. This makes expert revision, which our approach is targeted towards, even more important.

By calculating the measures presented in this section, we are able to analyze the training data set and extract potential labeling errors using only the trained classifier.
In our visual error correction approach, we make use of the suggested error characterization and treat these three error types differently, both, by calculating specialized error measures, and employing tailored visual guidance systems.

\subsection{Workflow Integration}

As we exploit a pre-trained classifier for error detection, a few considerations need to be made in order to integrate our approach into a standard classification workflow.
Before analyzing the data set, the classifier needs to be trained.
Here the classifier and the training process do not need to be altered at all.
The user can then reinspect misclassified samples based on our proposed visual guidance.
Additionally, if the number of data points to be reinspected is too small, experienced users can employ strict regularization or early stopping if they intend to control the number of training samples to reinspect, as the classification accuracy directly influences this number.
To be able to use the classification results as guidance towards possible errors, we assume that the network still has enough correctly labeled data to learn from, and guide the user towards incorrect labels.
While this assumption is likely to be true for most scenarios, if the data set is too small or contains too much noise, our approach will not function anymore as it relies on the classification results of the neural network.

To then get an idea about which items should be inspected again, all samples in the data set are classified once using the trained classifier.
In a typical neural network setting, this would include training, test, and validation data, as all of them can contain errors.
It is important to note that no evaluation of the model or further training is done at this point, so the data-split or training setup is not corrupted in any way.
This way, each data point is assigned a probability distribution over all classes.
We then present only misclassified samples through our visual guidance approach which we introduce in the next section. This way, the user has to look at far fewer items than if they would have to inspect all data points again. Our evaluation shows that this approach works well even when a large number of incorrect labels are present (see~\Cref{sec::evaluation}).

\section{Visual Error Correction}\label{sec::correction}

\begin{table*}[t]
  \centering%
  \caption{User tasks involved when improving training data.
  The user has to first, detect potential errors, then try to reason them, before he/she can resolve them.
  The table shows how these tasks are completed for the three identified error types.}
  \label{tab:tasks}
\begin{tabular}{l | c | c | c}
  \textbf{} & \textbf{Class Interpretation Error} & \textbf{Instance Interpretation Error} & \textbf{Similarity Error} \\ 
  \hline
  \textbf{Detect} & Many samples misclassified from $a$ to $b$ & Samples confidently misclassified & Similar/ identical samples \\
  \hline
  \textbf{Reason} & Error or bad classifier performance? & Error or bad classifier performance? & Error or intentional?\\ 
  \hline
  \textbf{Resolve} & Reassign multiple labels & Reassign individual label & Remove item \\
\end{tabular}
\end{table*}

While obtaining potential error candidates, as described above, is essential for improving training data sets, only through visual guidance users can \emph{detect} potential errors, and \emph{reason} about them.
Our visual guidance approaches help to do this for all three error types that typically occur in labeling processes.
Once errors have been reasoned about, they can directly be \emph{resolved}.
Again, the visual correction of data points, which involves the user tasks of detecting, reasoning about, and resolving potential errors, should be in line with the error types we propose. This interplay of user tasks and error types is shown in~\Cref{tab:tasks}.

\subsection{Error Detection}

\begin{figure}[b]
 \centering 
 \includegraphics[width=\columnwidth]{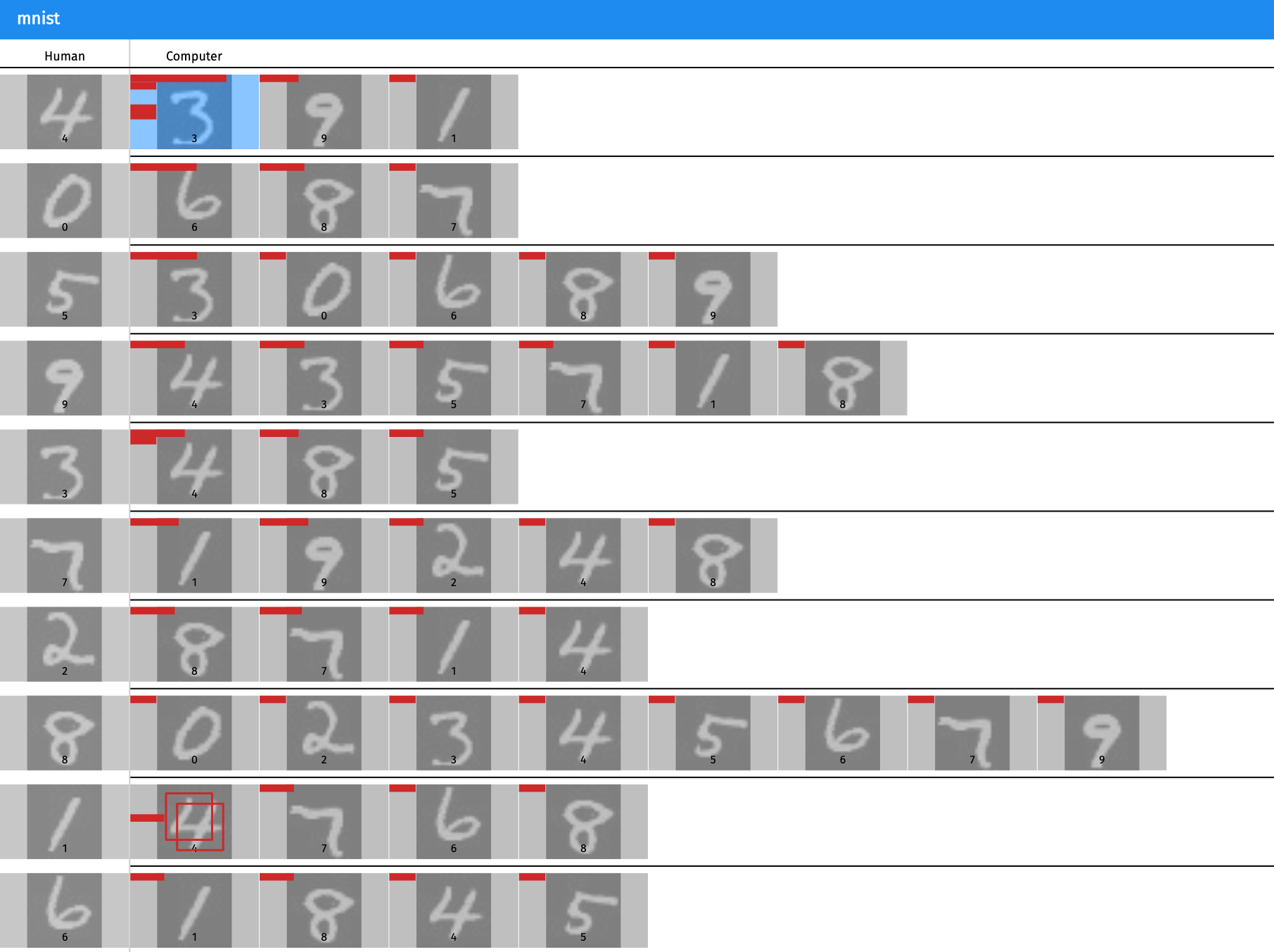}\hfill%
 \caption{The list view of classifications shows problematic lbl/cls combinations at a glance.
 The number of misclassifications for each cell is encoded in the blue background.
 The red horizontal bars in each cell show, how confidently the images have been misclassified as computed through~\Cref{eq:score}.
 Visual separation of rows makes clear, that this list should be read from left to right.
 On the left, one can see cells for correctly classified samples.}
 \label{fig::confusion_matrix}
\end{figure}

Through the error measures we propose, it is possible to support users through visual guidance to the most critical items in the data set.
For all three error types, users should have indications of which data points to review.
In~\Cref{sec::types}, we showed how candidates for these error types can be extracted from the data set based on classification results.
Thus, the user should be guided to lbl/cls pairs that contain a large number of misclassifications for Class Interpretation Errors.
For Instance Interpretation Errors, they should see which samples have been most confidently misclassified.
Additionally, users should be given an indication of where to find very similar images to be able to resolve Similarity Errors.
In the following, we present visual guidelines that support all of these requirements.
To give users an overview of those measures, we propose a visualization of the data set that contains information about the amount, probability distribution, and similarity score for each lbl/cls pair.
In line with our approach of guiding the user only to samples that the network misclassified, and thus might be labeled incorrectly, we only highlight misclassifications in this view, while correct classifications are depicted in the leftmost column. The resulting visualization can be seen in~\Cref{fig::confusion_matrix}.

We propose a visualization approach that employs a modified version of confusion matrices.
To search for possible Class Interpretation Errors, users need to be able to investigate lbl/cls combinations containing many misclassifications.
We support this requirement by sorting matrix cells based on the number of data points they contain, while the distribution of Instance Interpretation Scores is displayed within each cell.
We first sort by the number of misclassifications across different labels (rows), before sorting classification results within each label (columns).
This places the most critical classes at the top of this matrix.
Additionally, we omit cells that do not contain any items, which removes unnecessary clutter and makes the visualization more sparse.
In our implementation, we also highlight lbl/cls combinations with many misclassifications in blue, where the saturation of this color depends on the number of samples.
This guides the visual attention of users directly to these, most critical lbl/cls combinations.

To also embed the IIES-distribution of those misclassifications in this overview, which is helpful for spotting potential Instance Interpretation Errors, we propose to show this distribution using horizontal bar-charts within each list item.
Here, the y-position of the bars represents the IIES-distribution scaled from $1.0/num\_classes$ (lowest bar) to $1.0$ (top bar) while the length of the bars signals the number of items in an IIES-range.

The third user guidance system, which shows if similar items are present in a lbl/cls combination, is indicated by a duplicate icon within cells that contain highly similar data points.
With these visual indicators across the entire data set, this view serves as an overview that guides users to all three error types we defined in~\Cref{sec::types}.

Traditional approaches, such as confusion matrices~\cite{bilal2018convolutional,krizhevsky2009learning} or the confusion wheel~\cite{alsallakh2014visual}, which are commonly used to provide such an overview have major limitations for the task of spotting potential errors in the labeled data set.
Confusion matrices always require understanding and combining both, the label and the classification axis, which proved to be too complex for depicting the source and destination for misclassifications when presented to domain experts~\cite{ren2017squares}.
At the same time, most of the confusion wheels screen real estate is taken up by class overviews and it provides no clear entry point.
This renders both of these visual approaches suboptimal for guiding users to potential errors in the data set, which our approach is explicitly designed for.

\subsection{Error Reasoning}
When the user decides to inspect a potentially problematic lbl/cls combination, they naturally want to inspect individual data points and the distribution of data points in this subset of the data.
This way, they can reason about the potential errors to decide if they are problematic, and should be acted upon.
To inspect one such lbl/cls combination in detail, users select one of the items in our overview visualization.

\begin{figure*}[h!]
  \centering
  \setlength{\fboxsep}{0.0mm}
  \fbox{\includegraphics[height=5cm]{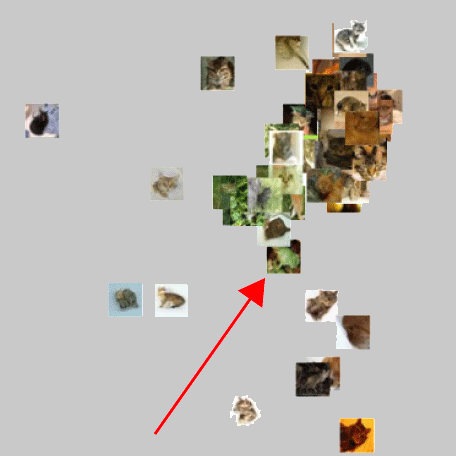}}
  \hfill
  \setlength{\fboxsep}{0.0mm}
  \fbox{\includegraphics[height=5cm]{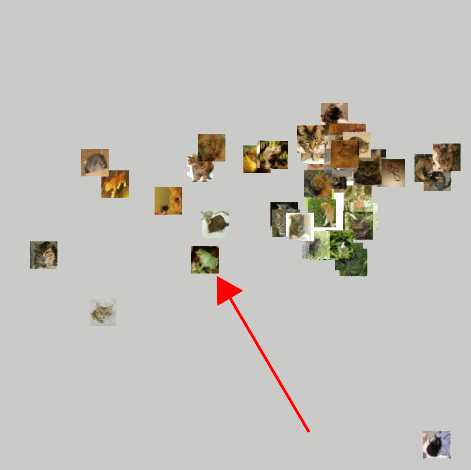}}
  \hfill
  \setlength{\fboxsep}{0.0mm}
  \fbox{\includegraphics[height=5cm]{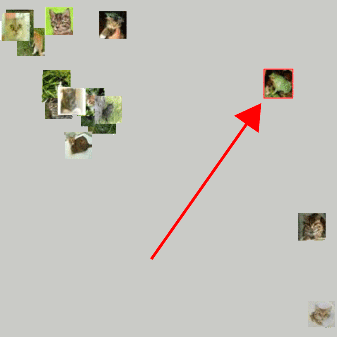}}
  \caption{
  UMAP~\cite{mcinnes2018umap} projection of the label cat and classification frog.
  One can see that dimensionality reduction helps to spot outliers in these lbl/cls combinations.
  The red arrows were added to indicate the position of the frog image.
  The three subsequent steps during interactive isolation of the frog wrongly labeled as cat show how after removing some data points, reprojecting the remaining data helps to isolate outliers.
  By iteratively removing outliers and through the nondeterministic nature of UMAP, the frog is embedded further away from the cats.
  (Images are from Cifar10, original resolution 32 by 32)\label{fig::find_frog}}
\end{figure*}

Reasoning about potential errors includes comparing samples, and extracting outliers as well as investigating similar samples for a lbl/cls combination.
Thus, we propose to guide the user by visualizing similarity-based embeddings of the selected lbl/cls combination.
Therefore, to inspect Instance Interpretation Errors, as well as Class Interpretation Errors, dimensionality reduction techniques that preserve high-dimensional relations are helpful.
If many similar items have been misclassified, users can quickly reason about potential Class Interpretation Errors as these items, which differ from plain misclassifications, will cluster when dimensionality reduction is applied.
On the other hand, outliers can be an indication for Instance Interpretation Errors, as can be seen in~\Cref{fig::find_frog}.
When dealing with images, we propose to use UMAP~\cite{mcinnes2018umap} to show clusters of data points, as well as outliers in this lbl/cls combination, which can be seen in~\Cref{fig::detail}.
Here, either direct image pixels can be used as projection features.
An even more sophisticated approach, which we used to generate these projections is, to use saliency visualizations of those images as a projection basis to also incorporate what the model looks for in these images.
While labeling errors will not always be projected as outliers, users can iteratively remove items from the visualizations by confirming or changing their labels, which eventually reveals label errors.
However, if there are few data points, or the user wishes to scan the data sequentially, there is also the option to switch to a grid-based view on the items.
To also support the inspection of Similarity Errors, the most similar images per /c combination should additionally be presented to the user.
In our implementation, those data points are shown below the projection-view.

Apart from showing data points with dimensionality reduction or sorted by similarity, their properties should also be inspectable in detail individually.
This can further help to decide upon whether a proposed error candidate was indeed labeled incorrectly.
Thus, in our proposed visualization approach, the final reasoning step on an individual data point level should be performed by selecting individual samples to view them in detail.
Additionally, for selected items, we show the probability distribution that stems from the classifier response to provide the user with another tool to reason about a potential labeling error.
In our implementation, enlarged images and classifier responses are shown on the right of the projection view (see~\Cref{fig::find_frog}).

While each of these visual guides is targeted towards satisfying a specific user-need, in combination, they provide the visual tools necessary to reason about the three error types we propose.

\subsection{Error Resolving}

\begin{figure}[b]
 \centering 
 \setlength{\fboxsep}{0.0mm}
  \includegraphics[width=\columnwidth]{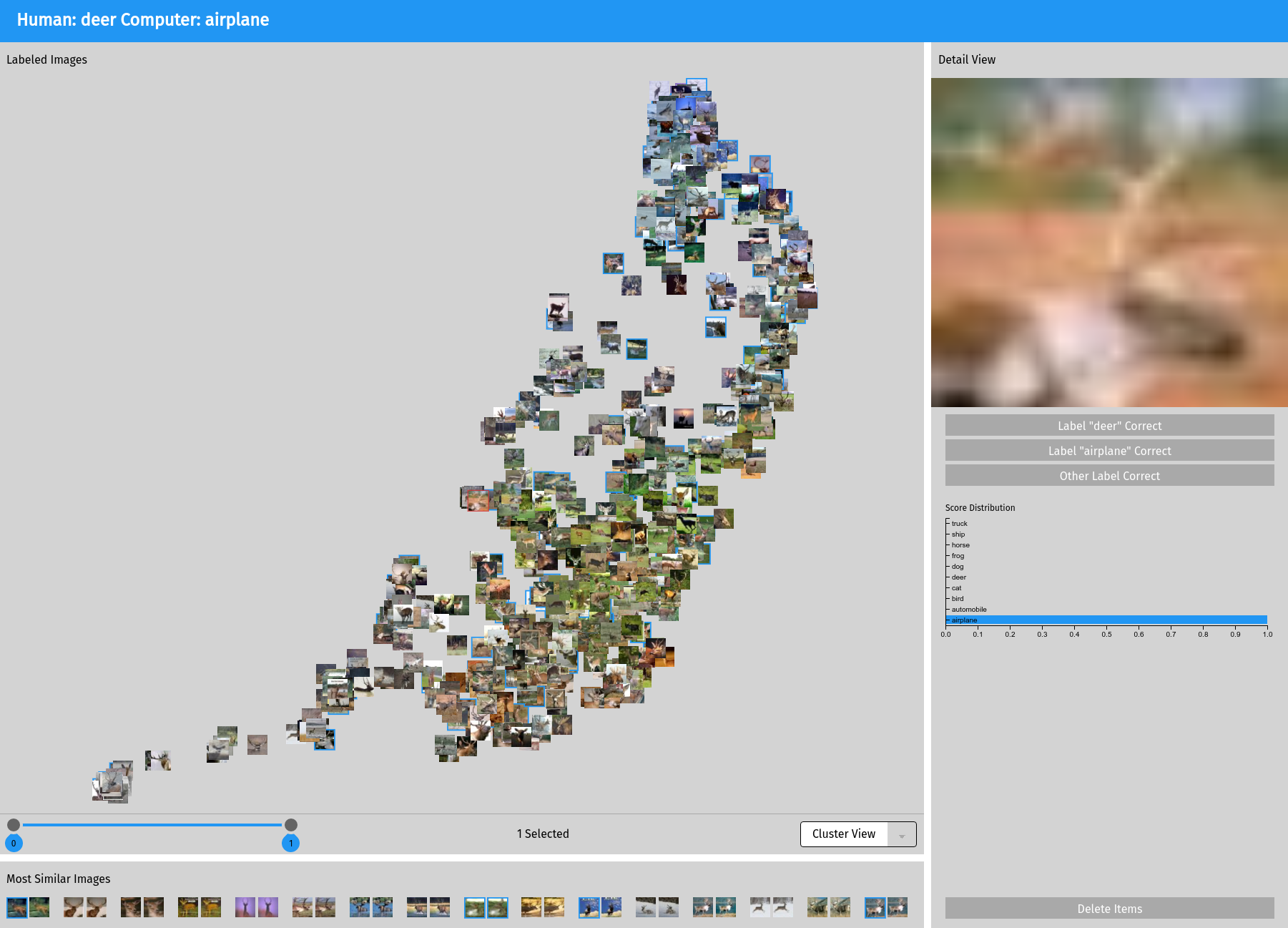}
 \caption{
   After gaining an overview of the classification results, the user can inspect the content of individual cells to analyze classification results in detail.
   Images are embedded by applying projection, e.g. UMAP.
   Filtering can be done by selecting IIES ranges.
   Once one or more images have been selected, the according probability distribution is visualized.
   Using the buttons on the right, users can change or confirm the label of the selected images.
   (Data set: Cifar10, resolution of images 32 by 32)\label{fig::detail}}
\end{figure}

The final step in our proposed iterative data-correction approach is resolving potential errors that have been found within the data set.
Once error candidates have been reasoned about, it is important to directly be able to resolve them.
This can mean assigning new labels, but also confirming labels that are correct to remove items from the error correction process.
For resolving Similarity Errors, data points should also be permanently removable from the data set.
To enable a correction, confirmation, and removal of labels for data points, we show actionable buttons on the lower right of the GUI (see~\Cref{fig::detail}).
Whenever data points are selected and subsequently resolved using these buttons, all visualizations are updated as resolved data points are removed from all guidance measure calculations and visualizations.
The effect of this can be seen in~\Cref{fig::find_frog}.
Thus, by resolving error candidates, users can interactively process the visualizations and work their way through the data set until all error candidates are resolved, and thus removed from the guidance approach.

After one iteration of data-correction has been completed, users can reiterate and restart the process by training a new classifier on the partially cleaned data set (see~\Cref{fig::process_new}).
With training a new classifier, proposed error candidates may change, and new error candidates can be inspected.
For subsequent iterations, our proposed measure calculation and user guidance can thus be kept as is, with the exception that all previously relabeled, removed, or confirmed data points are not included in the guidance system anymore, as they have already been resolved.

In our approach, users are guided to confident misclassifications, large quantities of misclassifications, and almost equal images through a data set overview, which helps to investigate potential errors.
To reason about error candidates, clustering mechanisms and outlier visualization are of great help.
It is also essential to directly be able to act upon inspected items to remove them from the process.
Through the translation of the three user tasks of detecting, reasoning about, and resolving potential labeling errors into our visualization guidelines, this approach can be implemented to fit any classifier as well as data type to be cleaned.
Thus, our approach enables a user-centered data cleaning that utilizes the trained classifier to propose error candidates.
The proposed visual design directly follows the principles of our approach to resolve the error types we introduced in~\Cref{sec::types}, and obeys to the user tasks we defined for the visual correction process.
Our implementation along with the user-study which we present in~\Cref{sec::evaluation} shows, that our concepts are applicable to network-supported data-cleanup, and could be adopted in many application scenarios.

\section{Evaluation}\label{sec::evaluation}

To test the proposed approach, we implemented a web-based application that realizes the proposed visualizations, and focuses on image data in combination with CNNs as classifiers.
The general idea of using the classifier as a guide to potential labeling errors is, however, not limited to such data or classification algorithms.
The following will present both, the application of our approach to renowned data sets, as well as a user study that tests the applicability of our approach.

\subsection{Analyzing Benchmark Data Sets}

Using our approach, we were able to spot errors in well-known machine-learning benchmark data sets.
Here, we analyzed both, the Cifar10~\cite{krizhevsky2009learning}, and MNIST~\cite{lecun-mnisthandwrittendigit-2010} data sets.

\noindent\textbf{MNIST.}
The MNIST data set~\cite{lecun-mnisthandwrittendigit-2010} is one of the most popular machine learning benchmark data sets.
It contains greyscale images of handwritten digits from zero to nine with a size of 28 by 28 pixels.
We used a simple architecture for training a CNN on that data set.
It consisted of two convolutional layers, each followed by a max-pooling layer.
For obtaining classification results on top of this embedding, one dense layer was used, followed by a dropout layer and the final softmax layer.
Our classifier reached an accuracy of 99.3 percent.
To review the data, we then inspected label classification pairs marked as suspicious in the overview visualization.
Since only 0.7 percent of the data set was misclassified, our visualization allowed us to only look at these images as potential errors.
Thus, instead of looking at all 70,000 images in a file explorer, we had to look at only 490 misclassified images through a guided process.
\\
When looking at the classes seven and one, some samples are almost impossible to distinguish while being from different classes.
This can be seen in~\Cref{fig:sevenone}.
Here, different cultural interpretations of said classes might lead to Class Interpretation Errors.
We found that the US-American versus European writing style of these digits might introduce problems to this data set.
We also discovered individual instances that are mislabeled in the MNIST data set.
~\Cref{fig::found} shows a data point that clearly shows a three, but was labeled as a five.
More of such examples can be found in our supplementary material.

\begin{figure}[t]
 \centering 
 \includegraphics[width=\columnwidth]{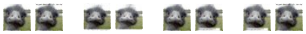}\hfill%
 \caption{
 At the bottom of our in-detail visualization, we show pairs of similar images.
 The user can then decide whether these should stay in the labeled data set (e.\,g.\, in cases where data augmentation is used) or if they should be removed (in case of unwanted duplicates).
 The images show five similar images of a bird discovered in the Cifar10 data set (original resolution 32 by 32).\label{fig::multiplebirds}}
\end{figure}

\noindent\textbf{Cifar10.}
The Cifar10 data set~\cite{krizhevsky2009learning} consists of 32 by 32 pixel colored images from ten different classes.
The model used for training on this data set was built by two blocks, each containing two convolutional layers followed by a pooling and a dropout layer.
This was then followed up by two dense layers each also preceding a dropout layer, before the final softmax classification layer was added.
With this intentionally simplistic network, we reached an accuracy of 77.13 percent, which is representative of real-world training scenarios on new, domain-specific data sets.
Even with the classification comparably low accuracy we reached, we only had to look at 22.87 percent of the data.
\\
As can be seen in~\Cref{fig::find_frog}, for Cifar10, we were able to spot an image that was incorrectly labeled as cat, while showing a frog.
When performing an in-detail inspection of the lbl/cls combination of the label cat and the classification frog, we found this incorrectly labeled image by iteratively removing outliers from the embedding visualization.
Additionally, we found a set of very similar bird images as shown in~\autoref{fig::multiplebirds}.
While this is not a clear error in the data set, having multiple highly similar images of an ostrich in this data set is at least debatable.

Our approach is generally targeted towards domain-experts that get their data labeled and then train a classifier on that data or use online services such as AutoML~\cite{automl2019} for training classifiers.
Here, these control mechanisms are even more important, as data quality can be worse than in benchmark datasets.
However, the fact that we were even able to find errors in two of the most popular benchmark data sets in the machine learning community shows how important approaches as the one we propose are.

\subsection{Qualitative Evaluation}

Based on our implementation, we additionally conducted a qualitative study to test the applicability of our approach.
In our user study, 10 participants had to find and resolve errors in a labeled data set.
Participants were recruited in a university setting, whereby out of the 10 participants, only two had experience with neural networks and none of them had seen or heard of our approach before.
This shows, that no ML background is needed to use our visualization guidelines.

To generate a setup in which we could recruit participants in a university setting while still reflecting a real-world scenario, where data-experts would correct their noisy data set using our approach, we chose to use the MNIST data set in our study.
This dataset requires no prior knowledge to review, as it consists of hand-drawn digits, which anyone can identify.
To be able to verify which items have been changed by a participant, we corrupted the data set by introducing five errors of each type.
For Class Interpretation Errors, we changed 1,400 images from nine to six, 700 images from one to four, 700 images from three to one, 350 images from eight to two and 175 images from seven to three.
For Instance Interpretation Errors, we changed the labels of five images from different classes.
With this, we tried to reflect real-world scenarios, where CIEs would introduce many more incorrect labels than IIEs.
Similarity Errors were introduced by duplicating five images.
In this study, we told the participants to remove all duplicates, as reasoning about if they are actually harmful could not be done in this setting.
In total, we introduced 3,330 mislabeled images and five duplicates.

We then trained on this data set and visualized the results using our implementation.
The classification accuracy for this manipulated data set was at 94.37 percent, hence, participants were only presented the 5.63 percent that were misclassified.
This equals to about 4,000 out of the 70,000 images.
We provided a short introduction of about 10 minutes which showed our idea for data-cleanup and explained the task, which was to resolve as many errors as possible in 15 minutes.
We then let them use the approach we propose in this paper to resolve all errors they spotted.


With our similarity guidance, all participants were able to resolve all duplicates.
We mainly attribute this to our visually prominent similarity indicators in the data set overview, and the fact, that the most similar items in a lbl/cls combination are shown separately when inspecting such combinations in detail.
On average, every participant changed the labels of 2,902 images, of which only 27.5 were incorrectly changed.
They thus managed to bring the number of incorrect labels down by 85.65 percent on average.
This is a reduction to 477 errors from 3,330 after only one iteration of our approach.
We then used the corrected data sets to train the classifier once again for each participant.
On average, the validation accuracy rose to 99.05 percent, which shows the enormous impact of such data-cleanup.
This shows the applicability of our approach to cleaning noisy labeled datasets.

Looking at the images that we initially considered as incorrectly changed also provided an interesting insight.
When investigating them, we found that some of them seemed to be mislabeled in the original data set.
The participants thus found new errors in the well-established MNIST data set by using our approach.
Examples of these errors are included in the supplementary material.

To also evaluate the usability of our techniques, we asked the participants to rate the helpfulness of our approach.
They had to rate the helpfulness of the visualizations from one, not helpful at all, to five, helped a lot, all of them rated the visualizations between four and five, with an average of 4.4.
When asked what they found most helpful, most of them said the overview guidance approaches were helpful for spotting errors in the data set.
Some additionally mentioned that it is also essential to be able to inspect individual samples for resolving errors.
When asked what was bad and could be improved, many said that the latency was a problem.
This, however, was a problem specific to the study setup and not to our approach perse.

As all participants were able to improve the data set by a large margin and thus greatly improve classification accuracy, this study shows that our proposed approach can, in fact, be a valuable tool to clean up labeled data.
Also, as our participants stated, our guidance system helps users focus on critical training samples which greatly reduces samples that need to be reinspected.

\section{Limitations}

Currently, the approach we present within this work is limited to classification problems.
For other problems, different error measures, as well as visual guidance systems, would have to be invented, which remains an open research question.
Additionally, the error types we present within this paper cannot be applied outside the domain of classification problems.
While our approach is model-agnostic and does not depend on the data that is used, the exemplar implementation we provide is focused on image-data in combination with CNNs.
We propose three types of errors, which our analysis of labeling processes suggests are most common.
However, one could think of other error cases, for example, if a labeler assigns completely random labels to all images.
We did not include such error cases, as most of them could be filtered by traditional quality assurance methods.
Nontheless, investigating and handling other potential labeling errors remains an open challenge.
Also, while matrix views are a common metaphor for getting an overview of classification results for a data set, and our proposed matrix is even more condensed than others, it cannot scale indefinitely.
We tested our approach with data sets containing up to more than 20 classes.
A data set with 22 different classes containing animal skull X-Ray images, can be seen in our supplementary material.
Yet, for data sets that contain even more classes, matrix views are not optimal.
In this case, users would have to look at a subset of classes rather than viewing the whole class-pool right away.
However, this is a general research question and is not only tied to our approach.

\section{Conclusion}

After introducing the problems that mislabeled training data for classification algorithms bring with them, we formulate a novel categorization of error types that typically occur in labeling settings for classification tasks.
While there are other approaches that aim at improving noisy labels in training data, ours introduces the concept of using the trained classifier as a support for resolving these three different error types.
The proposed visual correction approach can be performed at any point in the lifetime of a training data set, and permanently and reliably improves training data sets after the labeling process has been finished.
Contrary to other approaches, our visual error correction tightly couples automated approaches with user interaction to ensure data quality.
To model this visual correction approach, we define the user-tasks of first, detecting errors, then, reasoning about them, and finally resolving them, which users typically perform for cleaning up data sets.
Our method fits especially well into the context of crowdsourced data-labels.
With the ongoing automation of data acquisition, as well as classifier training, we imagine such data-cleanup techniques to be picked up in these contexts.
Our approach could be a candidate to be plugged in directly into services such as AutoML~\cite{automl2019}, where labels and classifiers can be obtained automatically, and correctly labeled data is crucial.

\section*{Acknowledgments}

This work was funded by the Carl-Zeiss-Scholarship for PhD students.

\bibliographystyle{eg-alpha-doi}
\bibliography{uccgcctd}
\end{document}